%% file: main.tex
\documentclass[conference]{IEEEtran}
\IEEEoverridecommandlockouts
\pdfoutput=1
\usepackage{cite}
\usepackage{amsmath,amssymb,amsfonts}
\usepackage{graphicx}
\usepackage{textcomp}
\usepackage{xcolor}
\usepackage[a4paper, total={184mm,239mm}]{geometry}
\def\BibTeX{{\rm B\kern-.05em{\sc i\kern-.025em b}\kern-.08em
    T\kern-.1667em\lower.7ex\hbox{E}\kern-.125emX}}

\usepackage{booktabs} 
\usepackage{gensymb}
\usepackage[english]{babel}
\usepackage{bm}
\usepackage[capitalise]{cleveref}
\usepackage{subcaption}
\usepackage{pifont}
\usepackage{makecell}
\usepackage{multirow}
\usepackage{multicol}
\usepackage{algorithm}
\usepackage{algpseudocode}

\usepackage{threeparttable}
\usepackage{booktabs}  
\usepackage{pifont}    
\newcommand{\cmark}{\ding{51}}
\newcommand{\xmark}{\ding{55}}

\makeatletter
\newcommand{\smallbullet}{} 
\DeclareRobustCommand\smallbullet{%
  \mathord{\mathpalette\smallbullet@{0.7}}%
}
\newcommand{\smallbullet@}[2]{%
  \vcenter{\hbox{\scalebox{#2}{$\m@th#1\bullet$}}}%
}
\makeatother

\begin{document}

\makeatletter
\newcommand{\linebreakand}{%
  \end{@IEEEauthorhalign}
  \hfill\mbox{}\par
  \mbox{}\hfill\begin{@IEEEauthorhalign}
}
\makeatother


\title{Diffuse to Detect: Generative Diffusion Models for Unsupervised IC Anomaly Detection
}
\author{
\IEEEauthorblockN{
  Yuxuan Yin\IEEEauthorrefmark{2}, 
  Chen He\IEEEauthorrefmark{3}, 
  Todd Jacobs\IEEEauthorrefmark{3}, 
  Jialei He\IEEEauthorrefmark{3}, 
  Boxun Xu\IEEEauthorrefmark{2}, 
  Robert Jin\IEEEauthorrefmark{3}, 
  Peng Li\IEEEauthorrefmark{2}\\
\IEEEauthorblockA{\IEEEauthorrefmark{2}Department of Electrical and Computer Engineering, University of California Santa Barbara, CA, USA}
\IEEEauthorblockA{\IEEEauthorrefmark{3}Automotive Processing, NXP Semiconductors, TX, USA \\}
\IEEEauthorblockA{\{y\_yin, boxunxu, lip\}@ucsb.edu, \{chen.he, todd.jacobs, jialei.he\_1, robert.jin\}@nxp.com}}
}


\maketitle

\begin{abstract}

Latent defect screening is challenged by extremely low failure rates, high-dimensional test data, and absence of labeled anomalies. We propose the first unsupervised anomaly detection framework incorporating a Diffusion Transformer. Raw test measurements are first compressed by an autoencoder, then reshaped into a structured token sequence enriched with sinusoidal and per-device wafer-position embeddings. Anomaly scores are derived from the noise-prediction error over mid-range diffusion timesteps, enabling fast wafer-scale screening without any labeled defects or manual feature engineering. Our approach achieves state-of-the-art performance on industrial 16nm IC test data under extreme class imbalance, offering interpretable failure localization through latent-space reconstruction residuals. 


\end{abstract}


\section{Introduction}

Semiconductor manufacturing relies on extensive electrical testing to screen defective devices before they reach end customers. As process nodes shrink and device complexity grows, the volume and dimensionality of test data expand correspondingly: modern production test programs generate thousands of measurement features per device, spanning leakage currents, propagation delays, voltage margins, and application-specific stress responses. Anomaly detection is applied using these parametric measurements to identify defective parts, which is critical to detect latent defects.

Latent defects are defects that can pass all functional test criteria yet harbor physical imperfections that will manifest as field failures under operating stress, for example, resistive shorts or resistive opens \cite{WLS}. Such latent defects are typically the roadblocks for automotive products to reach Zero Defect quality, as well as the root cause of the Silent Data Corruption (SDC) issue \cite{SDC}, which has been one of the most difficult challenges facing high-performance computing companies in recent years. Screening them is fundamentally difficult for three compounding reasons. First, latent defective devices are extremely rare. In mature processes, failure rates are measured in parts per million: in our industrial dataset, fewer than 10 latent outliers appear among approximately 6{,}000 nominally passing devices. This extreme class imbalance renders supervised learning approaches impractical---collecting sufficient labeled defect data is both costly and time-consuming, and classifiers trained on such skewed distributions tend to collapse to the majority class. Second, the test feature space is high-dimensional as there are typically thousands of parametric measurements during production test, introducing the curse of dimensionality and making distance- or density-based anomaly detection unreliable without careful feature selection. Third, the tolerance for screening error is extremely tight. Semiconductor yield loss is directly tied to economic viability: missed defects increase field return rates and warranty costs, while over-screening discards good dies and erodes yield. An effective screening system must operate at a precisely controlled false positive rate while maximizing recall of the rare anomalous units---a balance that na\"ive thresholding or generic anomaly detectors routinely fail to achieve.

The key observation that motivates our approach is that latent defects, despite passing functional tests, \emph{deviate from the distribution of normal devices} in the space of parametric test measurements. A device with a gate oxide weak spot may pass timing tests under nominal conditions yet exhibit subtly elevated leakage currents or anomalous delay margins that collectively fall outside the learned manifold of healthy silicon. This distributional deviation provides the signal for unsupervised detection, without requiring any defect labels.

Generative diffusion models~\cite{ho2020ddpm} are well suited to exploit this signal. Trained exclusively on the abundant healthy-device population, a diffusion model learns the joint distribution of normal test responses across all measurement dimensions. At inference, a device is scored by how poorly the model can reconstruct its test measurements---anomalous devices, lying outside the training distribution, induce elevated reconstruction error. This paradigm requires no labeled anomalies, scales naturally to large datasets, and sidesteps the manual feature engineering that has historically bottlenecked semiconductor quality pipelines. A further practical advantage is \emph{interpretability}: by examining the reconstruction residuals at each step of the reverse diffusion process, test engineers gain direct insight into which measurement regions deviate most strongly from normal behavior. This information for root-cause failure analysis that purely discriminative models cannot provide.

Prior diffusion-based anomaly detection has focused almost on image data~\cite{wyatt2022anodddpm, zhang2023diffusionad}, where spatial patch structure naturally maps onto transformer token sequences. The closest tabular counterpart, TabDDPM~\cite{kotelnikov2023tabddpm}, applies a ResNet-based denoising MLP directly in the raw feature space, treating the measurement vector as a flat, unordered input with no structural inductive bias. This design is adequate for low-dimensional tabular data but struggles with the regime of IC test data, where the raw feature space is both high-dimensional and semantically heterogeneous.

We address this gap with \textbf{Diffuse to Detect}, a fully unsupervised anomaly detection framework designed for high-dimensional IC test data. Our approach departs from TabDDPM in two fundamental ways. First, rather than diffusing in raw feature space, we learn a compact latent representation via an MLP encoder and perform diffusion in this lower-dimensional latent space. Second, we replace the flat ResNet denoiser with a \emph{1D Diffusion Transformer} (DiT1D)~\cite{peebles2022dit} that treats the latent representation as a structured token sequence, enabling the self-attention mechanism to capture compound inter-feature correlations. We further incorporate a per-device die positional embedding derived from the physical wafer coordinates of each device, accounting for the systematic spatial variation in measurements that is characteristic of wafer-level test data.

The contributions of this paper are as follows:
\begin{itemize}
    \item We introduce the first diffusion model-based framework for unsupervised anomaly detection on IC electrical test data, entirely eliminating the need for manual feature engineering.

    \item We introduce a two-level positional encoding scheme combining fixed sinusoidal token-position embeddings with a learned, per-device gated MLP die embedding, encoding both the semantic structure of the test flow and the spatial geometry of the wafer.

    \item We show that our approach outperforms established unsupervised baselines on industrial IC test datasets under the extreme class imbalance ($<\!0.12\%$ outlier rate) characteristic of mature semiconductor processes.
\end{itemize}

\input{sec/related}

\input{sec/method}

\input{sec/expr}

\section{Conclusion and Discussion}
We presented Diffuse to Detect, a fully unsupervised anomaly detection framework for high-dimensional IC test data. Raw parametric measurements are compressed by an autoencoder and reshaped into a structured latent token sequence, which a 1D Diffusion Transformer learns to denoise using healthy-silicon data exclusively. Two-level positional encodings capture both the semantic ordering of the test flow and the spatial geometry of the wafer, while dual scoring modes support both high-throughput production screening and expert failure analysis. Experiments on an industrial dataset with extreme class imbalance demonstrate competitive detection performance against classical and deep learning baselines, without any labeled defects or manual feature engineering. Latent-space reconstruction residuals provide interpretable anomaly localization that translates directly into actionable root-cause information for test engineers. 

Furture work can explore how to apply our framework for root-cause failure analysis of anormaly parametric features or test programs. For instance,  we can employ
a DDPM-style reconstruction score~\cite{wyatt2022anodddpm}. Starting
from the encoded latent $\mathbf{Z}_0$ without adding forward noise, we
run $T_{\text{rec}}$ steps of the learned reverse process:

\begin{equation}
  \mathbf{Z}_{t-1} = \boldsymbol{\mu}_\theta(\mathbf{Z}_t, t)
  + \sqrt{\tilde{\beta}_t}\,\boldsymbol{\xi},
  \quad \boldsymbol{\xi} \sim \mathcal{N}(\mathbf{0}, \mathbf{I}),
\end{equation}

yielding a reconstruction $\hat{\mathbf{Z}}_0$. The reconstructed latent is then decoded
back to the original measurement space via the decoder
$g_\phi$:

\begin{equation}
  \hat{\mathbf{x}} = g_\phi\!\left(
    \text{flatten}(\hat{\mathbf{Z}}_0)
  \right) \in \mathbb{R}^{F},
  \label{eq:decode}
\end{equation}

where $\text{flatten}$ collapses the $C \times L$ latent tensor to
$\mathbb{R}^{D_r}$ before decoding. The per-feature reconstruction
residual is then computed directly in the original measurement space:

\begin{equation}
  r_f = \left( x_f - \hat{x}_f \right)^2,
  \quad f = 1, \ldots, F,
  \label{eq:feature_residual}
\end{equation}

yielding a residual vector $\mathbf{r} = [r_1, \ldots, r_F] \in
\mathbb{R}^F$ that assigns a deviation score to every individual test
measurement.

Since features are produced by a structured sequence of $P$ test
programs $\{\mathcal{T}_1, \ldots, \mathcal{T}_P\}$, each generating a
contiguous block of $f_p$ features, the per-feature residuals can be
aggregated to a per-program anomaly score:

\begin{equation}
  s_p = \frac{1}{f_p} \sum_{f \in \mathcal{T}_p} r_f,
  \quad p = 1, \ldots, P,
  \label{eq:program_score}
\end{equation}

producing a compact anomaly profile $\mathbf{s} = [s_1, \ldots, s_P]
\in \mathbb{R}^P$ indexed by test program. This profile can be
visualized as a heatmap over the test flow, directly identifying which
test programs deviate most strongly from the healthy-silicon
distribution. A device with elevated $s_p$ for programs associated with
leakage measurement, for example, points toward a specific failure
mechanism without requiring any further analysis. Crucially, all
quantities are computed in the native units of the original test
measurements, making the output directly interpretable by test engineers
without knowledge of the latent representation.

\section*{Acknowledgment}
This work is supported by an NXP Long Term University (LTU) grant and the National Science Foundation under Grant No. 1956313.
\bibliographystyle{IEEEtran}
\bibliography{IEEEabrv, ref}

\end{document}

%% file: sec/related.tex
\section{Related Work}
\label{sec:related}

\subsection{Anomaly Detection in IC Testing}

Anomaly detection in semiconductor manufacturing and IC testing has been studied extensively, yet the dominant paradigm remains heavily reliant on engineered features and supervised or semi-supervised signal. Classical approaches apply statistical process control (SPC) techniques, such as multivariate control charts and principal component analysis, to hand-selected test parameters, flagging devices that fall outside learned normal operating bounds~\cite{montgomery2009spc, gu2020outlier}. While interpretable and computationally lightweight, these methods require significant domain expertise to define relevant features and thresholds, and do not generalize across product generations or process changes.

Machine learning methods have progressively supplemented rule-based approaches. Ensemble outlier detectors such as Isolation Forest~\cite{liu2008iforest} and one-class support vector machines~\cite{scholkopf2001ocsvm} have been applied to parametric test vectors, offering improved sensitivity to multivariate anomaly patterns. However, these methods scale poorly to the high-dimensional feature spaces produced by modern test programs, and remain dependent on curated feature inputs. Deep learning approaches, including autoencoders and variational autoencoders applied to equipment sensor data and electrical test outputs, have demonstrated improved representation learning without explicit feature engineering~\cite{liao2020stalad}. Most notably, TRACE-GPT~\cite{kim2023tracegpt} proposed a GPT-based generative pre-training framework adapted from natural language processing to model sequential semiconductor manufacturing sensor signals for unsupervised fault detection, demonstrating the promise of sequence-aware generative models in this domain. However, TRACE-GPT operates on continuous equipment sensor time series rather than the structured tabular outputs of parametric test programs, and employs an autoregressive generation objective rather than the diffusion-based anomaly scoring central to our approach.

Wafer bin map (WBM) analysis represents a parallel line of work in which spatial die-level pass/fail patterns are analyzed for defect fingerprinting~\cite{nakazawa2018wbm}. Diffusion models have recently been applied to WBM data for unknown defect pattern detection~\cite{moon2024wigdm}, leveraging reconstruction error on spatial image representations. Our work is fundamentally distinct: we operate on the raw continuous-valued parametric test measurements produced per device, not on spatial binary pass/fail maps, and we target the detection of subtle analog deviations in high-dimensional feature space rather than spatial clustering of hard failures.

\begin{table}[t]
  \centering
  \caption{Comparison of related anomaly detection methods.}
  \label{tab:related}
  \resizebox{\columnwidth}{!}{%
    \begin{tabular}{lccccc}
      \toprule
      \textbf{Method} & \textbf{Modality} & \textbf{Diffusion} & \textbf{Unsupervised} & \textbf{No Feat. Eng.} & \textbf{Token Localization} \\
      \midrule
      SPC ~\cite{montgomery2009spc}       & IC Test    & \xmark & \cmark & \xmark & \xmark \\
      Isolation Forest~\cite{liu2008iforest} & General & \xmark & \cmark & \xmark & \xmark \\
      TRACE-GPT~\cite{kim2023tracegpt}    & Sensor TS  & \xmark & \cmark & \cmark & \xmark \\
      AnoDDPM~\cite{wyatt2022anodddpm}    & Image      & \cmark & \cmark & \cmark & Pixel  \\
      TabDDPM~\cite{kotelnikov2023tabddpm}     & Tabular    & \cmark & \cmark & \cmark & \xmark \\
      DTE~\cite{livernoche2024dte}        & Tabular    & \xmark$^*$ & \cmark & \cmark & \xmark \\
      ImDiffusion~\cite{chen2023imdiffusion} & Time Series & \cmark & \cmark & \cmark & \xmark \\
      \midrule
      \textbf{Diffuse to Detect (Ours)}   & \textbf{IC Test} & \cmark & \cmark & \cmark & \textbf{Token} \\
      \bottomrule
    \end{tabular}%
  }
  \begin{tablenotes}
    \small
  \item $^*$ DTE estimates diffusion time via a surrogate network rather than running the generative denoising chain.
  \end{tablenotes}
\end{table}
\subsection{Diffusion Models for Anomaly Detection}

Denoising diffusion probabilistic models (DDPMs)~\cite{ho2020ddpm} learn a data distribution by training a neural network to reverse a gradual Gaussian noising process. The key property exploited for anomaly detection is that a model trained exclusively on normal data learns to reconstruct normal samples faithfully, while anomalous inputs yield elevated reconstruction error during the reverse process.

AnoDDPM~\cite{wyatt2022anodddpm} pioneered this approach for medical image anomaly detection, introducing simplex noise to control the spatial scale of detectable anomalies. Subsequent image-domain work has explored conditional diffusion for reconstruction~\cite{mousakhan2023ddad}, latent diffusion for scalability~\cite{graham2023ldm3d}, and iterative reconstruction-localization coupling~\cite{fucka2024transfusion}. These methods achieve strong performance on visual benchmarks such as MVTec~\cite{bergmann2019mvtec} and VisA, but their architectural assumptions---2D spatial patches, convolutional or vision transformer backbones, pixel-level anomaly maps---do not transfer to tabular test data.

For tabular data, diffusion-based anomaly detection remains largely unexplored. TabDDPM~\cite{kotelnikov2023tabddpm} is the first MLP diffusion model with residual connections for tabular data generation, and was performed with anomaly detection by reconstruction loss. However, TabDDPM treats tabular rows as flat, unordered vectors, making no use of any structural or semantic ordering among features. The Diffusion Time Estimation (DTE) method~\cite{livernoche2024dte} similarly operates on unordered tabular feature vectors, replacing the full reverse diffusion chain with a lightweight network trained to predict the noise level of a corrupted input, an efficient density proxy but not a true generative diffusion model, and without token level localization capability. As DTE does not support denoising instances, it can hardly be used for root-cause failure analysis.

For multivariate time series, ImDiffusion~\cite{chen2023imdiffusion} demonstrated that a transformer-backbone DDPM using imputation-based reconstruction, rather than full sequence reconstruction, yields superior anomaly detection by leveraging neighboring context. This insight is related to our use of positional structure, though ImDiffusion targets temporally ordered sensor streams with fixed, uniform spacing, whereas our tokens encode systematically wafer level variations.

\subsection{Diffusion Transformers}

The Diffusion Transformer (DiT)~\cite{peebles2022dit} replaced the convolutional U-Net backbone of standard DDPMs with a transformer operating on sequences of latent patches, demonstrating superior scalability and generation quality on image benchmarks. The transformer backbone brings two properties critical to our application: native support for variable-length token sequences with positional encoding, and per-token output resolution that enables token-level reconstruction scoring. While DiT was designed for image generation in a VAE latent space, its transformer denoising block is modality-agnostic. We adapt it to operate directly on latent test program token sequences, bypassing the autoencoder and patchification in favor of a 1D sequential tokenization aligned with the test flow structure of IC data.

%% file: sec/method.tex
\section{Methodology}
\label{sec:method}

\begin{figure}[t]
  \centering
  \includegraphics[width=\linewidth]{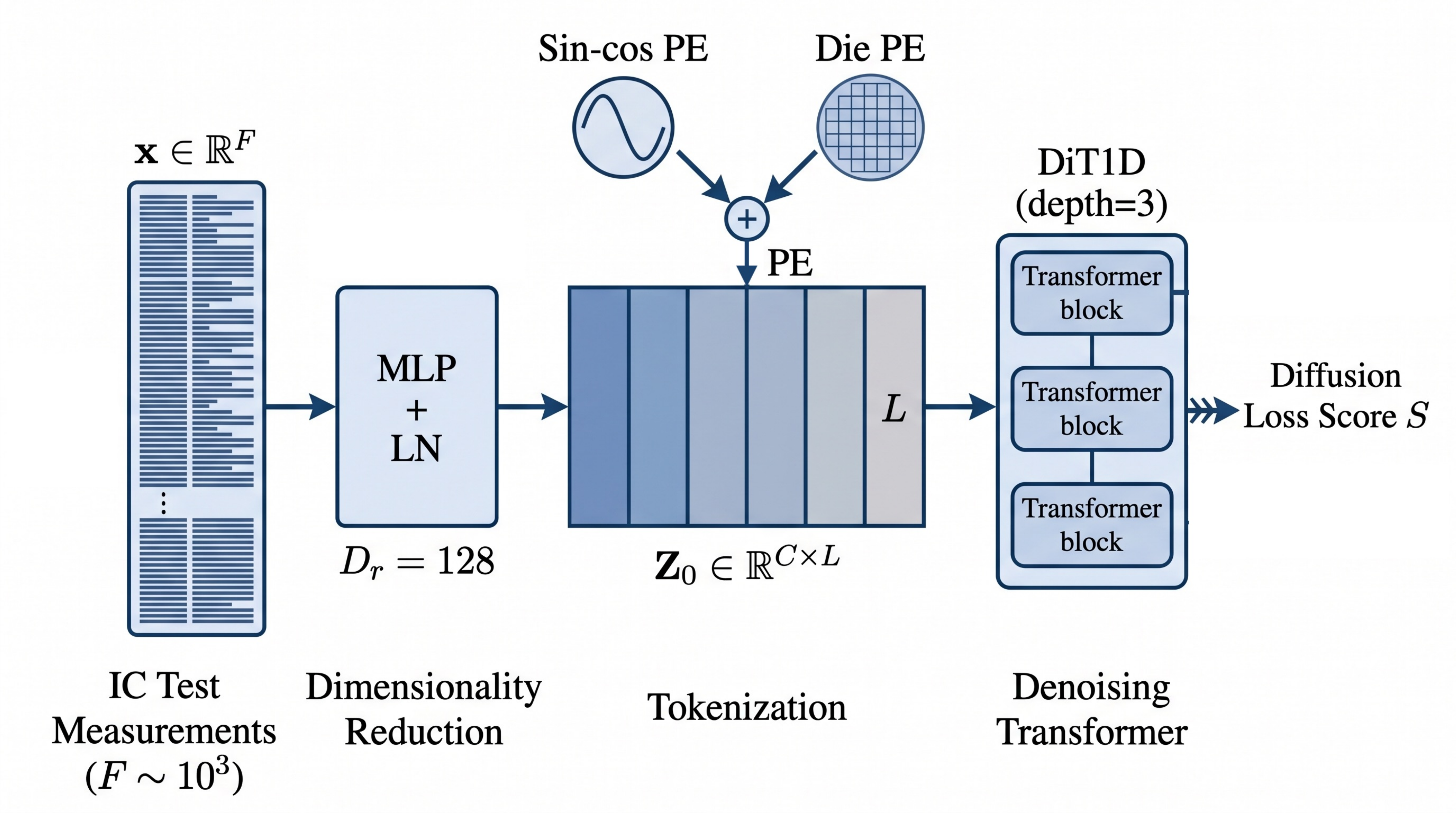}
  \caption{
    Overview of the proposed framework. A high-dimensional IC test
    measurement vector $\mathbf{x} \in \mathbb{R}^{F}$ is first reduced
    and tokenized into a latent sequence $\mathbf{Z}_0 \in
    \mathbb{R}^{C \times L}$, enriched with two levels of positional
    encoding, and processed by a 1D Diffusion Transformer (DiT1D) trained
    on healthy-silicon measurements. At inference time, diffusion loss is used as the anomaly score for high-throughput production screening.
  }
  \label{fig:overview}
\end{figure}

\subsection{Problem Formulation}
\label{sec:formulation}

Let $\mathcal{D} = \{\mathbf{x}^{(i)}\}_{i=1}^{N}$ denote a dataset of
$N$ IC devices, where each device is characterized by a high-dimensional
parametric test measurement vector
$\mathbf{x}^{(i)} \in \mathbb{R}^{F}$, with $F$ denoting the total
number of test features (in our setting, $F \sim 10^3$). These features
are produced by a structured sequence of $P$ test programs
$\{\mathcal{T}_1, \mathcal{T}_2, \ldots, \mathcal{T}_P\}$, where program
$\mathcal{T}_p$ generates a contiguous block of $f_p$ features, and
$\sum_{p=1}^{P} f_p = F$. 


We assume access to a training set $\mathcal{D}_{\text{train}} \subseteq
\mathcal{D}$ comprising predominantly healthy (passing) devices, collected
without anomaly labels. The goal is to learn the distribution of normal
test responses $p_{\theta}(\mathbf{x})$ and, at inference time, assign an
anomaly score $s(\mathbf{x}) \in \mathbb{R}$ to each device such that
anomalous devices receive significantly higher scores than healthy ones. An overview of the complete framework is illustrated in Fig.~\ref{fig:overview}.

\subsection{Dimensionality Reduction and Tokenization}
\label{sec:tokenization}

\begin{figure}[t]
  \centering
  \includegraphics[width=\linewidth]{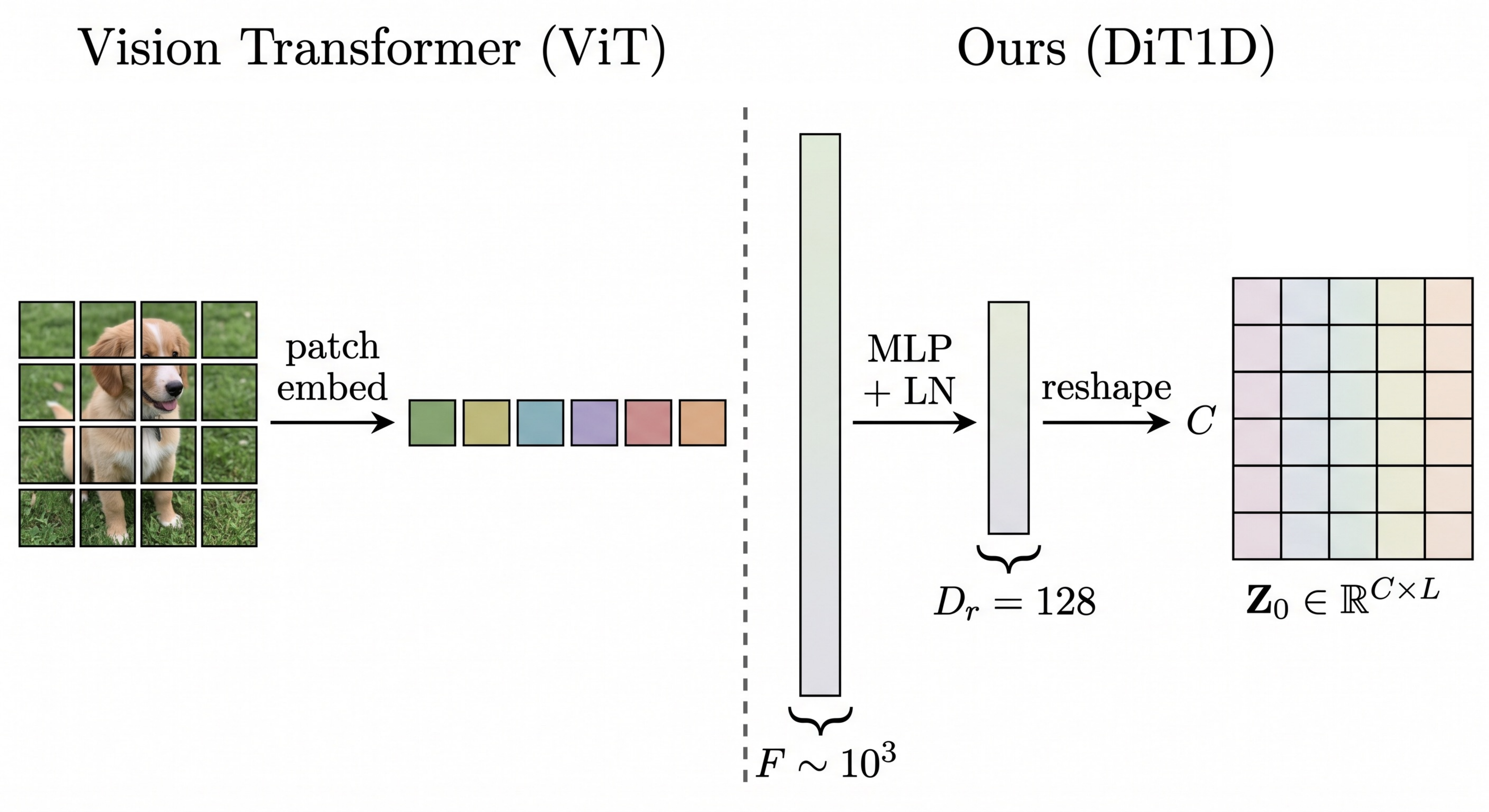}
  \caption{
    Dimensionality reduction and tokenization. The flat measurement
    vector $\mathbf{x} \in \mathbb{R}^{F}$ ($F \sim 10^3$) is
    projected by a two-layer MLP to $\mathbf{h} \in \mathbb{R}^{D_r}$
    ($D_r = 128$), then reshaped into a token sequence
    $\mathbf{Z}_0 \in \mathbb{R}^{C \times L}$. This is analogous to
    patch embedding in vision transformers, where an image is
    partitioned into spatial patches before entering the transformer.
  }
  \label{fig:tokenization}
\end{figure}

IC test measurement vectors are high-dimensional: each device is
characterized by $F \sim 10^3$ features spanning parametric test programs. Feeding such vectors directly into a
transformer would be computationally expensive and statistically
ill-conditioned, as the feature space far exceeds the number of
meaningful degrees of freedom in typical silicon responses. We therefore
decouple the pipeline into two sequential steps: (i) dimensionality
reduction to a compact representation, and (ii) tokenization of that
representation into a sequence suitable for the 1D DiT. This process is
illustrated in Fig.~\ref{fig:tokenization}.

\paragraph{Dimensionality Reduction}
A two-layer MLP with a SiLU activation reduces $\mathbf{x}$ to a
fixed-width representation:

\begin{equation}
  \mathbf{h} = \text{LN}\!\left(
    \mathbf{W}_2\,\sigma\!\left(\mathbf{W}_1 \mathbf{x} + \mathbf{b}_1\right)
    + \mathbf{b}_2
  \right),
  \quad \mathbf{h} \in \mathbb{R}^{D_r},
  \label{eq:reduction}
\end{equation}

where $\sigma$ denotes the SiLU nonlinearity and $D_r = 128 \ll F$
is the reduced dimension. The parameter-free Layer Normalization
$\text{LN}$ is applied after projection to enforce approximately
unit-variance statistics, which stabilizes the subsequent Gaussian
diffusion process~\cite{livernoche2024dte}. All input features are
standardized to zero mean and unit variance using statistics computed
over $\mathcal{D}_{\text{train}}$ prior to this step.

\paragraph{Tokenization}
The reduced representation $\mathbf{h}$ is reshaped into a 2D token
sequence that forms the input to the 1D DiT:

\begin{equation}
  \mathbf{Z}_0 = \text{reshape}(\mathbf{h};\; C, L)
  \in \mathbb{R}^{C \times L},
  \label{eq:tokenize}
\end{equation}

where $C$ is the number of latent channels and
$L = \lceil D_r / C \rceil$ is the sequence length (zero-padded if
$D_r$ is not divisible by $C$). This is directly analogous to patch
embedding in vision transformers~\cite{dosovitskiy2020vit}: just as a
ViT partitions an image into spatial patches to form a token sequence,
we partition the reduced measurement representation into $L$ tokens
along the sequence axis, each described by a $C$-dimensional feature
vector. The resulting sequence $\mathbf{Z}_0$ exposes the local
structure of the reduced representation to the self-attention mechanism
of the DiT, enabling the model to capture dependencies across different
regions of the latent space.

\subsection{Two-Level Positional Encoding}
\label{sec:pe}

\begin{figure}[t]
  \centering
  \includegraphics[width=\linewidth]{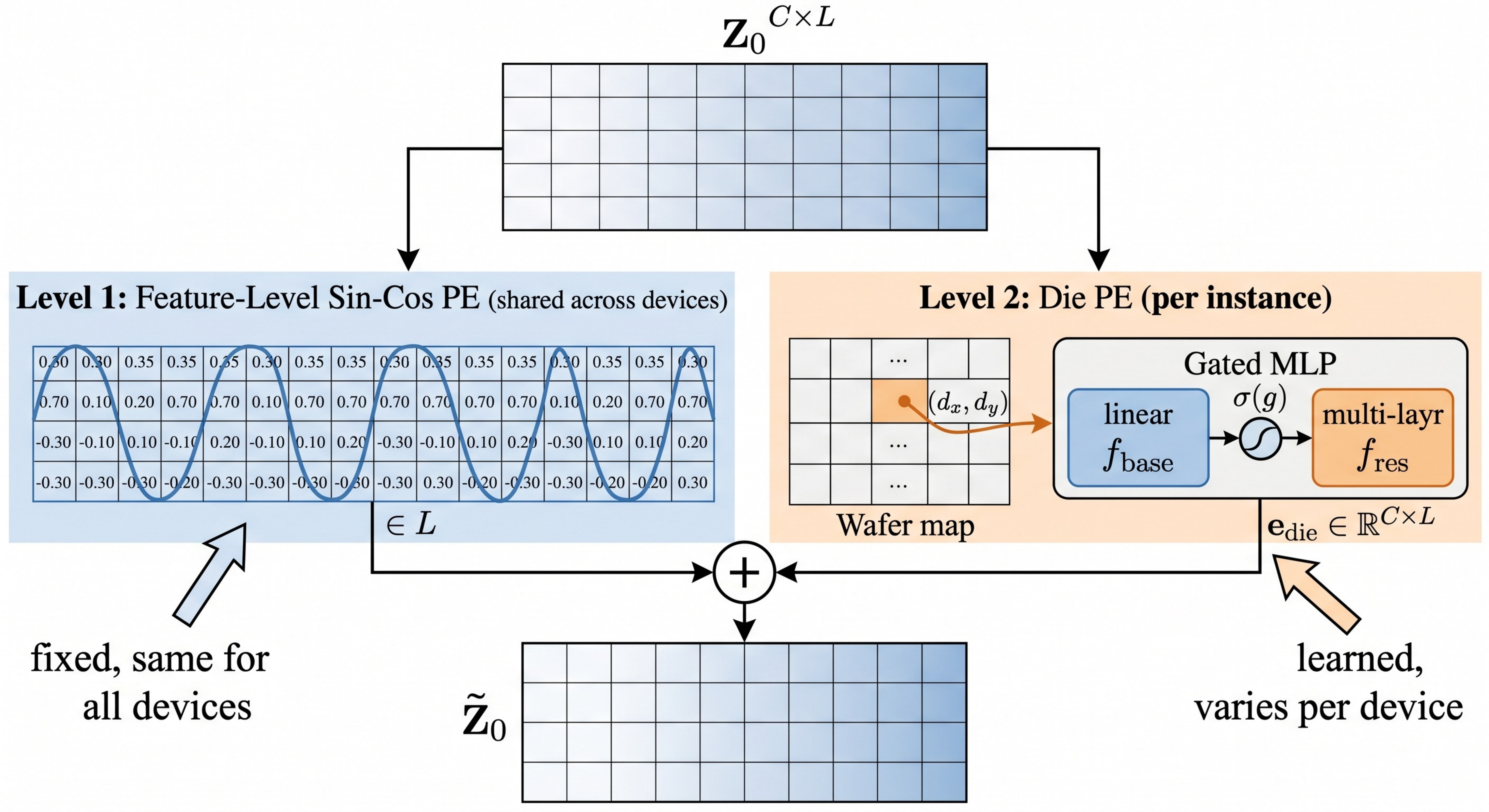}
  \caption{
    Two-level positional encoding. Feature-level:
    fixed sin-cos embeddings are added uniformly across all devices to
    encode token position within the latent sequence.
    Die-level: a gated MLP produces a per-instance
    embedding from the die's physical wafer coordinates $(d_x, d_y)$,
    capturing systematic spatial variation across the wafer.
  }
  \label{fig:pe}
\end{figure}

The token sequence $\mathbf{Z}_0$ receives positional information at two
distinct levels, encoding complementary structure in the data.
Fig.~\ref{fig:pe} illustrates both levels and how they are combined.

\paragraph{Feature-Level Sin-Cos Encoding}
To distinguish token positions within the latent sequence, we add a
fixed sinusoidal positional embedding over the $L$ sequence positions:

\begin{equation}
  \mathbf{Z}_0 \leftarrow \mathbf{Z}_0
  + \mathbf{E}_{\text{feat}} \in \mathbb{R}^{C \times L},
  \label{eq:feat_pe}
\end{equation}

where $\mathbf{E}_{\text{feat}}[\colon, \ell]$ is computed from the
standard 1D sinusoidal formula over positions $\ell = 0, \ldots, L-1$ \cite{vaswani2017attention}.
These embeddings provide a canonical and
consistent ordering of the latent token positions that is shared across
all devices.

\paragraph{Per-Instance Die Positional Embedding}
IC test data carries an important instance-level spatial covariate: the
physical $(d_x, d_y)$ location of each die on the wafer. Measurements
from dies at different wafer locations can exhibit systematic
distributional shifts (e.g., edge versus center effects), which are not
captured by the feature-level encoding.

We therefore add a per-sample \emph{die positional embedding}
$\mathbf{e}_{\text{die}} \in \mathbb{R}^{C \times L}$ derived from the
die coordinates:

\begin{equation}
  \mathbf{Z}_0 \leftarrow \mathbf{Z}_0
  + \mathbf{e}_{\text{die}}(d_x, d_y),
  \label{eq:die_pe}
\end{equation}

computed via a \emph{gated MLP} over a
coordinate feature vector that concatenates raw die coordinates with
standard per-axis sinusoidal embeddings~\cite{vaswani2017attention}:

\begin{equation}
  \mathbf{v} = \left[d_x,\; d_y,\;
    \text{sincos}(d_x),\; \text{sincos}(d_y)
  \right] \in \mathbb{R}^{2 + D_s},
  \label{eq:die_feat}
\end{equation}

where $D_s$ serves as the embedding dimension. The die positional embedding is then:

\begin{equation}
  \mathbf{e}_{\text{die}} = \text{reshape}\!\left(
    \mathbf{W}_{\text{out}}\!\left(
      f_{\text{base}}(\mathbf{v})
      + \sigma(g)\cdot f_{\text{res}}(\mathbf{v})
    \right);\; C, L\right),
  \label{eq:die_gated}
\end{equation}

where $f_{\text{base}}$ is a linear projection, $f_{\text{res}}$ is a
two-layer MLP residual branch, and $g$ is a learned scalar gate with
zero initialization, ensuring the embedding starts as a pure linear
function of the coordinates and acquires nonlinearity during training.
Unlike Level~1, this embedding varies per device: two dies at different
wafer locations receive different additive offsets to their token
sequences.

\subsection{Forward Diffusion on Token Sequences}
\label{sec:forward}

We adopt the Denoising Diffusion Probabilistic Model (DDPM)
framework~\cite{ho2020ddpm}. Given a clean token sequence
$\mathbf{Z}_0$, the forward process defines a Markov chain that
progressively corrupts the sequence by adding Gaussian noise over $T$
timesteps:

\begin{equation}
  q(\mathbf{Z}_t \mid \mathbf{Z}_{t-1})
  = \mathcal{N}\!\left(\mathbf{Z}_t;\,
    \sqrt{1 - \beta_t}\,\mathbf{Z}_{t-1},\;
  \beta_t \mathbf{I}\right),
\end{equation}

where $\{\beta_t\}_{t=1}^{T}$ is a fixed variance schedule. Using the
reparameterization $\bar{\alpha}_t = \prod_{s=1}^{t}(1-\beta_s)$, any
noisy sequence can be sampled in closed form:

\begin{equation}
  \mathbf{Z}_t = \sqrt{\bar{\alpha}_t}\,\mathbf{Z}_0
  + \sqrt{1 - \bar{\alpha}_t}\,\boldsymbol{\epsilon},
  \quad \boldsymbol{\epsilon} \sim \mathcal{N}(\mathbf{0}, \mathbf{I}).
  \label{eq:forward_closed}
\end{equation}

We use a cosine noise schedule as proposed in ~\cite{nichol2021improved}.

\subsection{Denoising Backbone: 1D Diffusion Transformer}
\label{sec:backbone}

\begin{figure}[t]
  \centering
  \includegraphics[width=\linewidth]{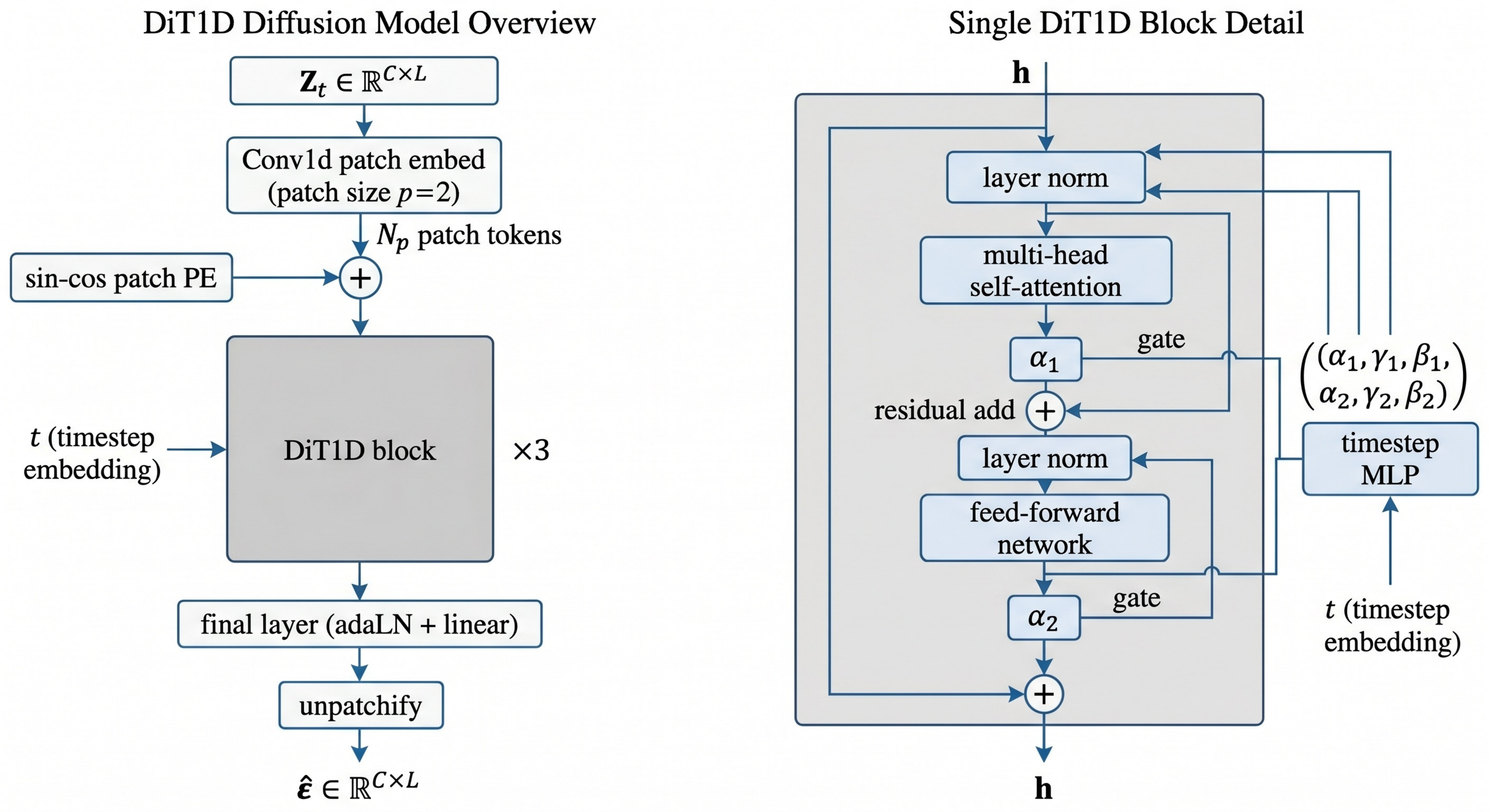}
  \caption{
    Architecture of the 1D Diffusion Transformer (DiT1D).
    Left: the full forward pass of diffusion model. The token sequence
    $\mathbf{Z}_t \in \mathbb{R}^{C \times L}$ is patchified by a
    \texttt{Conv1d} stem, enriched with patch-level sin-cos positional
    embeddings, passed through $D$ transformer blocks
    conditioned on the diffusion timestep $t$, and projected back to
    the original shape via the final layer and unpatchify.
    Right: detail of a single transformer block,
    showing timestep-conditioned scale-and-shift modulation applied to
    both the self-attention and feed-forward sub-layers.}
  \label{fig:dit1d}
\end{figure}

The reverse process learns to denoise $\mathbf{Z}_t$ back toward
$\mathbf{Z}_0$ by training a network
$\boldsymbol{\epsilon}_{\theta}(\mathbf{Z}_t, t)$ to predict the added
noise. We instantiate this as a \emph{1D Diffusion Transformer}
(DiT1D), adapted from the DiT architecture~\cite{peebles2022dit} for
sequential latent inputs. The architecture is shown in
Fig.~\ref{fig:dit1d}.

\paragraph{1D Patch Embedding}
The 2D patch embedding of the original DiT is replaced by a
\texttt{Conv1d}-based 1D patch embedding with patch size $p$, which
partitions the $L$-length token sequence into $\lfloor L/p \rfloor$
non-overlapping patches and projects each to the transformer hidden
dimension $d$:

\begin{equation}
  \mathbf{U} = \text{PatchEmbed1D}(\mathbf{Z}_t)
  + \mathbf{E}_{\text{patch}} \in \mathbb{R}^{N_p \times d},
\end{equation}

where $N_p = \lceil L/p \rceil$ is the number of patches and
$\mathbf{E}_{\text{patch}}$ is a fixed 1D sin-cos positional embedding
over patch positions.

\paragraph{Transformer Blocks}
Each transformer block applies multi-head self-attention followed by a
position-wise feed-forward network, with adaptive layer normalization conditioning on the diffusion timestep $t$:

\begin{equation}
  \mathbf{h} \leftarrow \mathbf{h}
  + \alpha_1 \cdot \text{Attn}\!\left(
  \text{LN}_{\gamma_1, \beta_1}(\mathbf{h})\right),
\end{equation}
\begin{equation}
  \mathbf{h} \leftarrow \mathbf{h}
  + \alpha_2 \cdot \text{FFN}\!\left(
  \text{LN}_{\gamma_2, \beta_2}(\mathbf{h})\right),
\end{equation}

where $(\alpha_1, \gamma_1, \beta_1, \alpha_2, \gamma_2, \beta_2)$ are
predicted by a small MLP from the sinusoidal embedding of $t$, following
the adaLN-Zero initialization of~\cite{peebles2022dit}.

\paragraph{Final Layer and Unpatchify}
The final layer applies adaLN modulation and projects each patch token
back to $p \cdot C$ dimensions. The output is unpatchified to recover
the original shape $\mathbb{R}^{C \times L}$, yielding the predicted
noise $\boldsymbol{\epsilon}_\theta(\mathbf{Z}_t, t)$.


\subsection{Training Objective}
\label{sec:training}

The model is trained on $\mathcal{D}_{\text{train}}$ using the standard
DDPM simplified objective~\cite{ho2020ddpm}, which minimizes the
expected mean squared error between predicted and actual noise:

\begin{equation}
  \mathcal{L} = \mathbb{E}_{t,\,\mathbf{Z}_0,\,\boldsymbol{\epsilon}}
  \left[
    \left\|
    \boldsymbol{\epsilon}
    - \boldsymbol{\epsilon}_{\theta}(\mathbf{Z}_t, t)
    \right\|^2
  \right],
  \label{eq:loss}
\end{equation}

where $t$ is sampled uniformly from $\{1, \ldots, T\}$ and
$\mathbf{Z}_t$ is computed via Eq.~\eqref{eq:forward_closed}. By
training exclusively on healthy-silicon devices, the model learns the
joint distribution of normal parametric test responses.
Anomalous devices, which lie outside this learned distribution, will
induce elevated reconstruction errors during inference.

\subsection{Anomaly Scoring}
\label{sec:inference}

\begin{figure}[t]
  \centering
  \includegraphics[width=\linewidth]{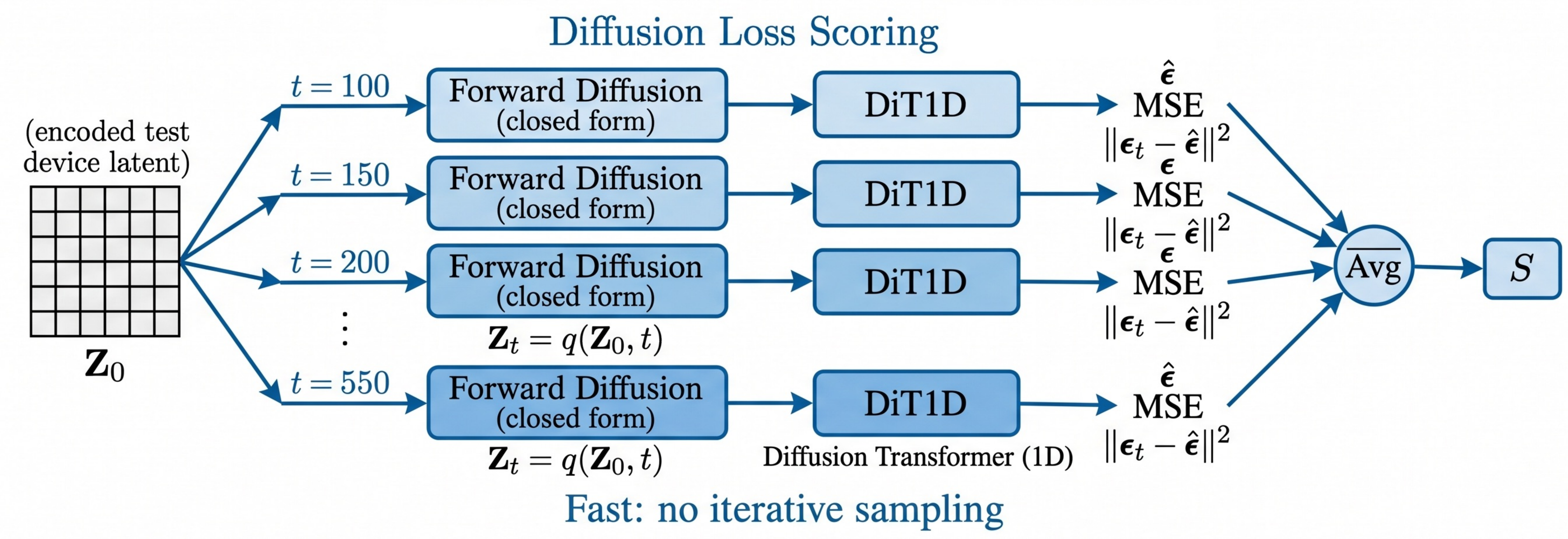}
  \caption{
    Anomaly scoring modes in our method. It evaluates the noise-prediction error over a fixed set of mid-range timesteps via forward diffusion.
  }
  \label{fig:scoring}
\end{figure}

For high-throughput production screening, we score devices via the
noise-prediction loss averaged over a fixed set of mid-range diffusion
timesteps, as illustrated in Fig.~\ref{fig:scoring}.. Given the latent $\mathbf{Z}_0$ obtained by encoding a test device
$\mathbf{x}$, we sample noisy versions at each evaluation timestep
using the closed-form forward process (Eq.~\ref{eq:forward_closed}) and
query the denoising network as anomaly score $S(\mathbf{x})$:

\begin{equation}
  S(\mathbf{x}) = \frac{1}{|\mathcal{T}_{\text{eval}}|}
  \sum_{t \in \mathcal{T}_{\text{eval}}}
  \left\|
  \boldsymbol{\epsilon}
  - \boldsymbol{\epsilon}_{\theta}(\mathbf{Z}_t, t)
  \right\|^2,
  \label{eq:diffusion_loss_score}
\end{equation}

where $\mathcal{T}_{\text{eval}} = \{t_{\text{start}},\,
t_{\text{start}} + \Delta t,\, \ldots,\, t_{\text{end}} - \Delta t\}$
is a fixed uniform grid over mid-range timesteps from 
$t_{\text{start}}$ to $t_{\text{end}}$ with step $\Delta t$.
This requires only $|\mathcal{T}_{\text{eval}}|$ forward passes through
the network with no iterative sampling, making it efficient for
large-scale wafer-level screening.

%% file: sec/expr.tex
\section{Experiments}
\label{sec:experiments}

\input{tab/dataset_split}

\subsection{Dataset and preprocessing}
\label{sec:dataset}

Our experiments use data collected from a 16\,nm automotive chip
product across two release versions. Each dataset is a table of one row
per (lot, wafer, die) test outcome, carrying a binary health label,
spatial identifiers (\texttt{lot\_key}, \texttt{wf\_key},
\texttt{die\_x}, \texttt{die\_y}), and a large set of parametric
measurements. Table~\ref{tab:dataset-splits} summarizes the resulting
dataset statistics after preprocessing.

\paragraph{Feature selection}
We select a task-specific feature subset by matching column names to a
regular expression (e.g.\ targeting parametric measurement columns),
ensuring the model never observes excluded test programs. Columns for
which the fraction of missing entries exceeds a fixed threshold
$r_{\text{na}}$ are removed. Any device that retains a missing value in
the remaining features is dropped, keeping only complete cases.

\paragraph{Normalization}
Rather than applying global standardization, we reduce wafer- and
lot-level offset via a within-wafer $z$-score: for every feature column
and each (\texttt{lot\_key}, \texttt{wf\_key}) group, we subtract the
group mean and divide by the group standard deviation, with a small
floor on the denominator to avoid degeneracy. Spatial identifiers
\texttt{die\_x} and \texttt{die\_y} are retained in their raw integer
grid for the die positional embedding and are never concatenated to the
normalized feature vector.

\paragraph{Train/test split}
We follow the standard unsupervised anomaly detection protocol. The
training set contains only healthy (label-normal) devices. The test set
pools the remaining healthy devices with all anomalous devices. Normal
devices are split equally between training and test (50\% each) using a
fixed random seed, yielding the sizes reported in
Table~\ref{tab:dataset-splits}. As shown, both datasets exhibit extreme
class imbalance, with anomaly rates of $0.22\%$ and $0.12\%$ for
Dataset~1 and Dataset~2 respectively---reflecting the rare-defect
regime characteristic of mature automotive semiconductor processes.

\subsection{Evaluation Metrics}
\label{sec:metrics}

Given the extreme class imbalance of our datasets, we report three
complementary metrics.

\paragraph{AUROC} (Area Under the Receiver Operating Characteristic
Curve) measures the probability that a randomly chosen anomalous device
receives a higher anomaly score than a randomly chosen normal device,
ranging from 0 to 1 with 0.5 indicating random performance. While
widely used, AUROC can be overly optimistic under severe class
imbalance~\cite{davis2006relationship}.

\paragraph{AUCPR} (Area Under the Precision-Recall Curve) is more
informative under extreme imbalance, as it focuses on the detector's
ability to retrieve anomalies at high precision. A random classifier
achieves an AUCPR equal to the anomaly rate ($\approx 0.22\%$ for
Dataset~1 and $\approx 0.12\%$ for Dataset~2), providing a meaningful
lower bound for comparison.

\paragraph{Recall@95\%~Yield} measures whether at least one confirmed
anomalous device is ranked above the decision threshold, set such that
$95\%$ of normal test devices are passed (i.e.\ $5\%$ are screened
out). This yield-constrained operating point directly reflects
production economics: tightening the threshold to increase recall comes
at the cost of yield loss, and a detector that fails to surface any
confirmed defect at this operating point provides no actionable value
regardless of its AUROC or AUCPR.

\subsection{Baselines}
\label{sec:baselines}

We compare against a comprehensive set of unsupervised anomaly detection
baselines spanning classical, deep learning, and diffusion-based methods.
Classical methods include Isolation Forest (IForest)~\cite{liu2008iforest},
One-Class SVM (OCSVM)~\cite{scholkopf2001ocsvm}, COPOD~\cite{li2020copod},
ECOD~\cite{li2022ecod}, Feature Bagging~\cite{lazarevic2005feature},
HBOS~\cite{goldstein2012hbos}, KNN~\cite{ramaswamy2000knn},
LODA~\cite{pevny2016loda}, LOF~\cite{breunig2000lof},
MCD~\cite{hardin2004mcd}, and PCA reconstruction error.
Deep learning-based approaches include
DAGMM~\cite{zong2018dagmm},
DROCC~\cite{goyal2020drocc},
GOAD~\cite{bergman2020goad},
ICL~\cite{shenkar2022icl},
PlanarFlow~\cite{rezende2015normalizing},
GANomaly~\cite{akcay2019ganomaly},
and SLAD~\cite{xu2023fascinating}.
Diffusion-based baselines include TabDDPM~\cite{kotelnikov2023tabddpm}, and two variants of
Diffusion Time Estimation: DTE-IG, and DTE-C\cite{livernoche2024dte},
which use inverse-gamma and, categorical distributions
respectively to estimate diffusion time as an anomaly score.
All baselines are implemented using
PyOD~\cite{zhao2019pyod} or their respective official codebases,
with hyperparameters set to their published defaults.

\subsection{Implementation Details}
\label{sec:impl}

The MLP encoder projects $F$-dimensional input features through a hidden
layer of dimension 128 with SiLU activation to a bottleneck of
$D_r = 128$, followed by parameter-free LayerNorm. The bottleneck is
reshaped into a token sequence of shape $C \times L$ with $C = 4$
channels and $L = \lceil 128 / 4 \rceil = 32$ positions. The DiT1D
denoiser uses a Conv1d patch embedding with patch size $p = 2$, yielding
$\lceil 32 / 2 \rceil = 16$ transformer tokens, and applies $T = 3$
adaLN-Zero transformer blocks with hidden dimension $d = 256$ and 4
attention heads. A symmetric MLP decoder (hidden dim 128, SiLU) maps the
bottleneck back to $F$ dimensions and is used only during the autoencoder
pretraining phase. The die positional embedding uses the gated MLP
variant with sinusoidal coordinate features of dimension 64 and a hidden
size of 256. The total parameter count of the full model
(encoder + DiT1D + decoder) is approximately 4M.

Training proceeds in two phases following the latent diffusion model style schedule. In the
first $K = 50$ epochs, only the MLP encoder and decoder are trained on
the MSE reconstruction loss, with the DiT1D denoiser frozen. In the
subsequent 200 epochs, the encoder is frozen and the DiT1D is trained on
the DDPM simplified noise-prediction objective with $T = 1{,}000$
diffusion steps and a cosine noise
schedule~\cite{nichol2021improved}. Both phases use the AdamW
optimizer~\cite{loshchilov2017decoupled} with learning rate $10^{-4}$ and
weight decay $5 \times 10^{-4}$. All experiments use a batch size of
2{,}048 and a fixed random seed of 42 for reproducibility.

Anomaly scores are computed via diffusion loss scoring
(Eq.~\ref{eq:diffusion_loss_score}) over a uniform grid of timesteps
$\mathcal{T}_{\text{eval}} = \{100, 150, \ldots, 550\}$, totaling 10
forward passes through the denoiser per device. The die positional
embedding is applied at both training and inference time using the raw
integer grid coordinates (\texttt{die\_x}, \texttt{die\_y}) recorded
for each device. All experiments are run on a single NVIDIA A100 80GB GPU.

\input{tab/main_small}

\input{tab/main_large}

\subsection{Main Results}
\label{sec:results}

Tables~\ref{tab:main-small} and~\ref{tab:main-large} report anomaly
detection performance on Dataset~1 and Dataset~2 respectively. Across
both datasets, DiT1D achieves the highest AUROC and Recall@95\%~Yield,
demonstrating consistent gains over all classical, deep learning, and
diffusion-based baselines.

\paragraph{Dataset~1} The
majority of classical and deep learning methods fail to surface a single
anomaly at the 95\% yield operating point, with AUROC values clustering
near chance (0.44--0.70). DiT1D achieves an AUROC of 0.771, surpassing
the second-best method LOF (0.716) by a substantial margin, and an
AUCPR of 0.0250---more than four times the second-best score of 0.0062
(DTE-IG). Critically, DiT1D recalls 3 out of 7 confirmed anomalies at
the 95\% yield threshold, while all competing methods surface at most
one. This large gap in AUCPR and recall confirms that DiT1D produces
well-calibrated anomaly scores concentrated on true positives, rather
than merely achieving a globally favorable ranking.

\paragraph{Dataset~2}
Dataset~2 is substantially larger than dataset 1. Classical
methods become more competitive in terms of recall, with LODA and HBOS
recalling up to 6 anomalies, while most deep learning methods degrade
significantly (AUROC 0.40--0.58). DiT1D achieves the highest AUROC
(0.639) and is the only method to recall all 7 anomalies present in
the test split at the 95\% yield operating point, suggesting that the
latent-space diffusion scoring is more robust to the increased
feature dimensionality than both classical density estimators and
raw-space deep learning baselines. While DTE-IG achieves the highest AUCPR
(0.0055) on this dataset, its recall of 5 confirms that a
favorable precision-recall curve does not guarantee recovery of the
rarest anomalies at a fixed yield constraint.

Across both datasets, vanilla TabDDPM consistently underperforms other
diffusion-based methods, confirming that diffusing in raw feature space
without structural inductive bias is insufficient for high-dimensional
IC test data. The consistent advantage of DiT1D over TabDDPM and DTE
variants validates our two key design choices: learning a compact latent
representation before diffusion, and using a transformer denoiser that
captures inter-token correlations across the measurement sequence.

\input{tab/ablation}

\subsection{Ablation Study}
\label{sec:ablation}

We ablate the three principal design choices of DiT1D on Dataset~1,
reporting performance as the decrease from the full model.
Table~\ref{tab:ablation} summarizes the results.

\paragraph{Die-level positional embedding}
Removing the gated die PE causes the largest drop in AUROC ($-0.234$),
confirming that wafer-spatial variation is a strong confounding factor
in high-dimensional IC test data. Without this embedding, the model
cannot distinguish systematic edge-versus-center measurement shifts
from genuine distributional anomalies, inflating the false positive
rate and suppressing recall.

\paragraph{Latent-space autoencoder}
Removing the MLP encoder and performing diffusion directly in the raw
feature space leads to the largest drop in both AUCPR ($-0.021$) and
Recalled ($-3$), reducing recall to zero at the 95\% yield threshold.
This confirms that compressing $F \sim 10^3$ features into a structured
latent representation before diffusion is critical: raw-space diffusion
is poorly conditioned in the high-dimensional IC test regime.

\paragraph{Transformer vs.\ MLP denoiser}
Replacing the DiT1D denoiser with a TabDDPM-style MLP backbone (while
retaining the latent-space encoder) degrades AUROC by $-0.079$ and
Recalled by $-2$. This isolates the contribution of the transformer's
self-attention mechanism: capturing inter-token correlations across the
latent sequence provides a consistent benefit over a position-agnostic
MLP operating on the same compressed representation.

\subsection{Effect of Model Depth}
\label{sec:depth}

Table~\ref{tab:num_blocks} reports performance as a function of the
number of DiT1D transformer blocks. A single block (depth~1) achieves
an AUROC of 0.700 but fails to recall any anomaly at the 95\% yield
threshold, suggesting insufficient capacity to model the joint
distribution of healthy silicon. Performance peaks at depth~3
(AUROC~0.771, AUCPR~0.025, Recalled~3) and degrades with a fourth
block (AUROC~0.754, AUCPR~0.009, Recalled~1), indicating mild
overfitting of the denoiser to the training distribution. We therefore
adopt depth~3 as our default configuration, which balances model
capacity against the dataset scale of $N \sim 3{,}000$ healthy training
devices.

\input{tab/num_blocks}

\subsection{Effect of Patch Size}
\label{sec:patchsize}

The patch size $p$ of the Conv1d stem controls the granularity of the
token sequence fed to the transformer: a smaller patch yields more
tokens and finer-grained attention, at the cost of a longer sequence.
Given a latent sequence of length $L = \lceil D_r / C \rceil = 32$
(with $D_r = 128$, $C = 4$), the number of transformer tokens is
$\lceil L / p \rceil$.

Table~\ref{tab:patchsize} reports results for patch sizes $p \in
\{2, 4, 8\}$, corresponding to 16, 8, and 4 tokens respectively.
Performance degrades sharply as patch size increases: $p = 4$ reduces
AUROC to 0.719 and Recalled to 2, while $p = 8$ collapses AUROC to
0.343 and fails to recall any anomaly. This trend indicates that
coarser tokenization discards fine-grained latent structure that is
informative for anomaly detection, and that the self-attention
mechanism benefits from a longer sequence to model dependencies across
the compressed measurement representation. We adopt $p = 2$ as the
default.

\begin{table}[t]
\centering
\caption{Effect of patch size on Dataset~1. Tokens denotes the number
of transformer input tokens after Conv1d patchification of the
$L\!=\!32$ latent sequence. Our default ($p\!=\!2$) is highlighted.}
\label{tab:patchsize}
\resizebox{\columnwidth}{!}{%
\begin{tabular}{ccccc}
\toprule
\textbf{Patch size} $p$ & \textbf{Tokens}
  & \textbf{AUROC} $\uparrow$
  & \textbf{AUCPR} $\uparrow$
  & \textbf{Recalled} $\uparrow$ \\
\midrule
2 & 16 & \textbf{0.771} & \textbf{0.025} & \textbf{3} \\
4 & 8 & 0.719          & 0.012          & 2           \\
8 &  4 & 0.343          & 0.011          & 0           \\
\bottomrule
\end{tabular}%
}
\end{table}

%% file: tab/dataset_split.tex
\begin{table}[t]
  \centering
  \caption{Dataset statistics after preprocessing.}
  \label{tab:dataset-splits}
  \begin{tabular}{@{}l r r r r@{}}
    \toprule
    Dataset & \# Features & \# Samples & \# Anomalies & Anomaly rate \\
    \midrule
    Dataset~1 & 1{,}158 & 6{,}255 &  7 & $0.22\%$ \\
    Dataset~2 & 3{,}034 & 69{,}009 & 41 & $0.12\%$ \\
    \bottomrule
  \end{tabular}
\end{table}

%% file: tab/main_small.tex
\begin{table}[t]
\centering
\caption{Anomaly detection performance on the IC parametric test dataset 1.
Recalled: number of confirmed anomalies detected
at $95\%$ yeild. Best result per metric is \textbf{bold};
second best is \underline{underlined}.}
\label{tab:main-small}
\resizebox{\columnwidth}{!}{%
\begin{tabular}{llccc}
\toprule
\textbf{Category} & \textbf{Method} & \textbf{AUROC} $\uparrow$
  & \textbf{AUCPR} $\uparrow$ & \textbf{Recalled} $\uparrow$ \\
\midrule
\multirow{11}{*}{Classical}
  & IForest~\cite{liu2008iforest}          & 0.550 & 0.0033 & 0 \\
  & OCSVM~\cite{scholkopf2001ocsvm}        & 0.631 & 0.0034 & 0 \\
  & COPOD~\cite{li2020copod}               & 0.508 & 0.0027 & 0 \\
  & ECOD~\cite{li2022ecod}                 & 0.464 & 0.0024 & 0 \\
  & FeatureBagging~\cite{lazarevic2005feature} & 0.699 & 0.0053 & 1 \\
  & HBOS~\cite{goldstein2012hbos}          & 0.474 & 0.0025 & 0 \\
  & KNN~\cite{ramaswamy2000knn}            & 0.610 & 0.0036 & 0 \\
  & LODA~\cite{pevny2016loda}              & 0.446 & 0.0025 & 0 \\
  & LOF~\cite{breunig2000lof}              & \underline{0.716} & 0.0056 & 1 \\
  & MCD~\cite{hardin2004mcd}               & 0.489 & 0.0024 & 0 \\
  & PCA                                    & 0.444 & 0.0023 & 0 \\
\midrule
\multirow{7}{*}{Deep Learning}
  & DAGMM~\cite{zong2018dagmm}             & 0.452 & 0.0024 & 0 \\
  & DROCC~\cite{goyal2020drocc}            & 0.670 & 0.0044 & 0 \\
  & GOAD~\cite{bergman2020goad}            & 0.601 & 0.0032 & 0 \\
  & ICL~\cite{shenkar2022icl}              & 0.673 & 0.0041 & 1 \\
  & PlanarFlow~\cite{rezende2015normalizing} & 0.467 & 0.0026 & 0 \\
  & GANomaly~\cite{akcay2019ganomaly}      & 0.645 & 0.0059 & 1 \\
  & SLAD~\cite{xu2023fascinating}          & 0.552 & 0.0029 & 0 \\
\midrule
\multirow{4}{*}{Diffusion-based}
  & TabDDPM~\cite{kotelnikov2023tabddpm}                 & 0.467 & 0.0025 & 0 \\
  & DTE-IG~\cite{livernoche2024dte}        & 0.563 & \underline{0.0062} & 1 \\
  & DTE-C~\cite{livernoche2024dte}         & 0.578 & 0.0031 & 0 \\
  & \textbf{DiT1D(Ours)}                    & \textbf{0.771} & \textbf{0.0250} & \textbf{3} \\
\bottomrule
\end{tabular}%
}
\end{table}

%% file: tab/main_large.tex
\begin{table}[t]
\centering
\caption{Anomaly detection performance on the second dataset.
Recalled: number of confirmed anomalies detected
at $95\%$ yeild. Best result per metric is \textbf{bold};
second best is \underline{underlined}.}
\label{tab:main-large}
\resizebox{\columnwidth}{!}{%
\begin{tabular}{llccc}
\toprule
\textbf{Category} & \textbf{Method} & \textbf{AUROC} $\uparrow$
  & \textbf{AUCPR} $\uparrow$ & \textbf{Recalled} $\uparrow$ \\
\midrule
\multirow{11}{*}{Classical}
  & IForest~\cite{liu2008iforest}          & 0.546 & \underline{0.0026} & 4 \\
  & OCSVM~\cite{scholkopf2001ocsvm}        & 0.595 & 0.0020 & 0 \\
  & COPOD~\cite{li2020copod}               & 0.525 & 0.0020 & 4 \\
  & ECOD~\cite{li2022ecod}                 & 0.562 & 0.0023 & 4 \\
  & FeatureBagging~\cite{lazarevic2005feature} & \underline{0.629} & 0.0026 & 3 \\
  & HBOS~\cite{goldstein2012hbos}          & 0.538 & 0.0025 & 5 \\
  & KNN~\cite{ramaswamy2000knn}            & 0.588 & \underline{0.0026} & 5 \\
  & LODA~\cite{pevny2016loda}              & 0.548 & 0.0021 & \underline{6} \\
  & LOF~\cite{breunig2000lof}              & 0.625 & 0.0024 & 3 \\
  & MCD~\cite{hardin2004mcd}               & 0.530 & 0.0015 & 3 \\
  & PCA                                    & 0.558 & 0.0021 & 4 \\
\midrule
\multirow{8}{*}{Deep Learning}
  & DAGMM~\cite{zong2018dagmm}             & 0.425 & 0.0011 & 2 \\
  & DROCC~\cite{goyal2020drocc}            & 0.431 & 0.0010 & 1 \\
  & GOAD~\cite{bergman2020goad}            & 0.399 & 0.0010 & 1 \\
  & PlanarFlow~\cite{rezende2015normalizing} & 0.580 & 0.0018 & 5 \\
  & GANomaly~\cite{akcay2019ganomaly}      & 0.426 & 0.0011 & 2 \\
  & SLAD~\cite{xu2023fascinating}          & 0.463 & 0.0012 & 2 \\
  & DIF                                    & 0.415 & 0.0010 & 1 \\
\midrule
\multirow{4}{*}{Diffusion-based}
  & TabDDPM~\cite{kotelnikov2023tabddpm}                 & 0.523 & 0.0015 & 2 \\
  & DTE-IG~\cite{livernoche2024dte}        & 0.545 & \textbf{0.0055} & 5 \\
  & DTE-C~\cite{livernoche2024dte}         & 0.503 & 0.0012 & 2 \\
  & \textbf{DiT1D (Ours)}                 & \textbf{0.639} & 0.0023 & \textbf{7} \\
\bottomrule
\end{tabular}%
}
\end{table}

%% file: tab/ablation.tex
\begin{table}[t]
\centering
\caption{Ablation study on dataset 1. Results shown as decrease ($\downarrow$) from \textbf{DiT1D (Ours)}.
A larger decrease indicates a more important component.}
\label{tab:ablation}
\resizebox{\columnwidth}{!}{%
\begin{tabular}{lccc}
\toprule
\textbf{Method} & $\Delta$\textbf{AUROC} $\downarrow$
  & $\Delta$\textbf{AUCPR} $\downarrow$ & $\Delta$\textbf{Recalled} $\downarrow$ \\
\midrule
w/o die-level PE   & $-0.234$ & $-0.018$ & $-1$ \\
w/o AutoEncoder    & $-0.124$ & $-0.021$ & $-3$ \\
w/ TabDDPM         & $-0.079$ & $-0.015$ & $-2$ \\
\bottomrule
\end{tabular}%
}
\end{table}

%% file: tab/num_blocks.tex
\begin{table}[t]
\centering
\caption{Effect of number of TabDiT blocks on dataset 1. Our chosen configuration (3 blocks) is highlighted.}
\label{tab:num_blocks}
\resizebox{\columnwidth}{!}{%
\begin{tabular}{lccc}
\toprule
\textbf{\# TabDiT Blocks} & \textbf{AUROC} $\uparrow$
  & \textbf{AUCPR} $\uparrow$ & \textbf{Recalled} $\uparrow$ \\
\midrule

1 block  & 0.700          & 0.012          & 0 \\
2 blocks & 0.769          & 0.013          & 2 \\
3 blocks & \textbf{0.771} & \textbf{0.025} & \textbf{3} \\
4 blocks & {0.754} & {0.009} & {1} \\
\bottomrule
\end{tabular}%
}
\end{table}